\def\year{2021}\relax
\documentclass[letterpaper]{article} 
\usepackage{aaai21}  
\usepackage{times}  
\usepackage{helvet} 
\usepackage{courier}  
\usepackage[hyphens]{url}  
\usepackage{graphicx} 
\usepackage{natbib}  
\usepackage{caption} 
\usepackage{multirow}

\title{A Context Aware Approach for Generating Natural Language Attacks}
\author {

        Rishabh Maheshwary,\textsuperscript{}
        Saket Maheshwary,\textsuperscript{}
        Vikram Pudi \textsuperscript{} \\
}
\affiliations {
    \textsuperscript{} Data Sciences and Analytics Center, Kohli Center on Intelligent Systems\\
International Institute of Information Technology, Hyderabad, India \\
    \{rishabh.maheshwary, saket.maheshwary\}@research.iiit.ac.in, vikram@iiit.ac.in
}

\begin{document}

\maketitle

\begin{abstract}
We study an important task of attacking natural language processing models in a black box setting. We propose an attack strategy that crafts semantically similar adversarial examples on text classification and entailment tasks. Our proposed attack finds candidate words by considering the information of both the original word and its surrounding context. It jointly leverages masked language modelling and next sentence prediction for context understanding.  In comparison to attacks proposed in prior literature, we are able to generate high quality adversarial examples that do significantly better both in terms of success rate and word perturbation percentage.
\end{abstract}
\section{Introduction}
Recent studies have shown evidence that deep neural networks (DNNs) are vulnerable to \emph{adversarial examples} --- inputs
generated by adding small perturbations to the original sample. Such perturbations are almost indistinguishable to humans but deceive DNNs, thus raising serious concerns about their utility in real world applications.
While prior work related to vision and speech have a variety of methods for crafting adversarial examples, it is still a challenging problem in natural language processing (NLP) because of $(1)$ the discrete nature of text, as replacing a single
word in a sentence can completely alter its meaning, and $(2)$ grammatical correctness and fluency. Existing black box attacks in NLP~\cite{jin2019bert,ren2019generating} generate adversarial examples using a two step process. $(1)$ It finds important words which highly impact the classification score of the target model. $(2)$ It replaces those words with a synonym generated either by using a lexical database such as WordNet or from the nearest neighbour in the counter-fitted embedding space. Though this method is effective in generating adversarial attacks, it has the following drawbacks. $(1)$ It takes only the word level similarity to find synonyms but does not consider the overall context surrounding the replaced word thus generating out of context synonyms which degrades the overall semantics. $(2)$ It yields unnatural and complex replacements which results in grammatically incorrect and non-fluent adversarial examples.

In this paper, we propose a black box attack that crafts high quality adversarial examples on text classification and entailment tasks. For each word to be replaced, our proposed attack generates candidate words using the influence of both the original word (to be replaced) as well as its surrounding context. The generated candidates $(1)$ fit well within the sentence thus retaining the overall semantics of the sentence, $(2)$ have the same meaning as that of the original word and $(3)$ are grammatically correct and fluent. We use BERT to generate candidates for each word to be replaced in the input. BERT is a masked language model trained on masked language modeling (MLM) --- predicts the $[MASK]$ word using both left and right context and next sentence prediction (NSP) --- predicts the next sentence given previous sentences separated by the $[SEP]$ token. For each word to be replaced, we substitute that word with a $[MASK]$ token to get a masked input. Then, we leverage the NSP and feed both the original input as well as the masked input separated by the $[SEP]$ to BERT.
\section{Proposed Approach}
Given a target model $\bf{F}: \mathcal{X}\rightarrow \mathcal{Y}$, that classifies an input text $\mathcal{X}$ to a set of class labels $\mathcal{Y}$. Our goal is to generate an adversarial text sequence $\mathcal{X_{ADV}}$ that is misclassified by $\bf{F}$ i.e. $\bf{F}(\mathcal{X}) \ne\bf{F}(\mathcal{X_{ADV}})$ and is semantically similar to $\mathcal{X}$.
We solve this problem using the following two steps\footnote{Code: \url{https://github.com/RishabhMaheshwary/contextattack}}:
\begin{table}[h!]
\centering
\begin{tabular}{|c|c|c|c|c|c|}
\hline
\textbf{Model} & \textbf{Orig\%} & \textbf{Attack} & \textbf{Acc\%} & \textbf{Pert\%} & \textbf{I\%}\\ \hline
{BERT} & \multirow{2}{*}{90.9} & TF & 13.6 & 6.1 & 0.5 \\
 (IMDB) & & Ours & \bf{9.5} & \bf{4.2} & \bf{0.38} \\ \cline{1-6} 
 & \multirow{4}{*}{89.8} & GA & 3.0 & 14.7 & 0.78 \\
  LSTM & & PWWS & 2.0 & 3.38 & 0.46 \\
 (IMDB) &  & TF & 0.3 & 5.1 & 0.53\\
  &  & Ours & \bf{0.3} & \bf{3.2} & \bf{0.32} \\ \hline
 {BERT} & \multirow{2}{*}{86.5} & TF & 11.5 & 16.7 & 1.26 \\
 (MR) &  & Ours & \bf{10.7} & \bf{16.3} & \bf{0.7} \\ \cline{1-6} 
  & \multirow{2}{*}{80.0} & TF & 3.1 & 14.9 & 1.04 \\
   {LSTM} &  & PWWS & 3.7 & 14.38 & 0.86 \\
  (MR) &  & Ours & \bf{2.1} & \bf{14.0} & \bf{0.7}\\ \hline
 {BERT} & \multirow{2}{*}{89.1} & TF & 4.0 & \bf{18.5} & 9.7 \\
 (SNLI) &  & Ours & \bf{3.6} & 18.9 & \bf{1.9} \\ \cline{1-6} 
  & \multirow{3}{*}{84.0} & GA & 30 & 23.3 & 9.9 \\
 {LSTM} & & TF & 3.5 & 18.0 & 7.7 \\
 (SNLI) &  & Ours & \bf{3.0} & \bf{17.0} & \bf{2.3} \\ \hline
\end{tabular}
\caption{Result comparison. Orig\% is the original accuracy, Acc\% is the after attack accuracy, Pert\% is the average perturbation rate, I\% is the grammatical error increase rate.}
\label{table:1}
\end{table}

{\textbf{{Word Ranking:}}} Given an input text $\mathcal{X} = \{x_1,x_2...x_n\}$, this step assigns high scores to words which significantly impact the final prediction of $\bf{F}$. To score a word $x_i$, this step removes $x_i$ from the input and queries $\bf{F}$ to observe the change in the classification score of the target class. This process is repeated for all of the words in $\mathcal{X}$. All the words are then sorted in descending order based on their score.

{\textbf{{Word Substitution:}}} Given the word $x_i$ and $\mathcal{M}$ the BERT masked language model, this step finds a word replacement for $x_i$ using the information of both the original word $x_i$ and its surrounding context. It consists of the following steps:

$(1)$ {{Candidate Generation:}} It replaces $x_i$ with a $[MASK]$ token to get a masked input $\mathcal{X}_{i} = \{x_1..x_{i-1},[MASK],x_{i+1}..x_n\}$. But, simply masking a word and using $\mathcal{M}$ to predict the masked word might generate candidate words that are not synonyms of $x_i$. For example, consider the sentence "I love this movie", if the word ``love'' is masked than $\mathcal{M}$ might predict words like ``hate'' and ``dislike'' with high probability which in turn will alter the overall semantics of the input. Therefore, to capture the influence of the original word $x_i$, we leverage NSP and feed both the original input $\mathcal{X}$ as well as the masked input $\mathcal{X}_i$ separated by a $[SEP]$ token to $\mathcal{M}$. Feeding the sentence pair ($\mathcal{X}$,$\mathcal{X}_i$) generates candidate words which not only fits the given context but also retain the meaning of the original word $x_i$. We take the top $\mathcal{K}$ predicted words for each word $x_i$ for substitution. Further, to ensure that the generated candidates are highly similar to the original word $x_i$ we apply counter-fitting and filter out candidates with cosine similarity less then the threshold.

$(2)$ {{{Final candidate selection:}}} Each remaining candidate is substituted for $x_i$ in $\mathcal{X}$ which results in a perturbed text sequence $\mathcal{X}^{'}_{i}$. Each perturbed sequence is checked for semantic similarity with $\mathcal{X}$ and sequences with threshold less than $\lambda$ are filtered out. The remaining perturbed sequences are fed to the target model $\bf{F}$. The perturbed text sequence which alters the final prediction of the target model is selected as the final adversarial example. In case the prediction does not change then the perturbed sequence which results in the least confidence score of the target class is selected and the above steps are repeated on the chosen perturbed sequence for the next ranked word.

\textbf{Experiments} We evaluate our proposed attack strategy across two state-of-the art target models ---  WordLSTM, BERT-base-uncased on three benchmark datasets --- IMDB, MR and SNLI. Note that the BERT masked language model $\mathcal{M}$ used to generate candidates is different from the BERT target model as the latter is fine tuned for each dataset. We use \emph{after attack accuracy} --- accuracy achieved on the generated adversarial examples, \emph{perturbation rate} --- the percentage of words substituted in an input and \emph{grammatical correctness} --- average grammatical error increase rate as our evaluation metrics. For all these three metrics, the lower the value, better the attack result. \begin{table}[h!]
\centering
\begin{tabular}{|l|l|}
\hline
\textbf{Type} & \textbf{Examples} \\ \hline
\textbf{Orig.} & Vile and tacky are best adjectives to describe it.\\
\textbf{TF} & Vile and tacky are best \textbf{qualifier} to describe it.\\
\textbf{Ours} & Vile and tacky are best \textbf{words} to describe it.\\ \hline
\textbf{Orig.} & This film is a portrait of grace in this world.\\
\textbf{TF} &  This film is a \textbf{spitting} of grace in this \textbf{universe}.\\
\textbf{Ours} & This film is a \textbf{sketch} of \textbf{beauty} in this world.\\ \hline
\end{tabular}
\caption{Demonstrates adversarial examples generated on BERT by TF and our attack. Substituted words are in bold.}
\label{table:2}
\end{table} We compare our attack with three state-of-the-art baselines, $(1)$  TextFooler (TF)~\cite{jin2019bert} $(2)$ PWWS~\cite{ren2019generating} $(3)$ Genetic Attack (GA)~\cite{alzantot2018generating}. We attack the target model on a test set of $1000$ examples, sampled from the test set of each dataset. To ensure fair comparison, these $1000$ examples are the same set of samples as used in TF and GA. We use Universal Sequence Encoder (USE) to measure the semantic similarity between the original and the adversarial example. We set the semantic similarity threshold ($\lambda$) to $0.8$ for classification and $0.6$ for entailment tasks. We take $30$ as the context window, and  set $\mathcal{K}$ to $50$.

\textbf{Attack Performance} As shown in Table $1$, we are able to achieve more than $90\%$ success rate across all target models and datasets. In comparison to TextFooler, on average our attack reduces the after attack accuracy and perturbation rate by $15\%$ and $14\%$ respectively across all target models and datasets. Further it reduces the grammatical error rate by at least $30\%$. Our attack also significantly outperforms other baselines on all evaluation metrics.\\
\section{Conclusion and Future Work}
We proposed an attack strategy to generate adversarial examples on classification and entailment tasks that outperforms baselines on multiple evaluation metrics across all target models and datasets. Although, our method generates fluent and grammatically correct adversarial examples, there is still scope for improvement. $(1)$ Some of the candidates generated by the candidate generation step are not synonyms of the original word. $(2)$ As our attack queries $\mathcal{M}$ for each word to be substituted, the attack generation process slows down. We leave these improvements for future work.  
\bibliography{context_attack.bib}
\end{document}